# Localization of License Plate Using Morphological Operations

` V.Karthikeyan and V.J.Vijayalakshmi

*Abstract-* It is believed that there are currently millions of vehicles on the roads worldwide. The over speed of vehicles ,theft of vehicles, disobeying traffic rules in public, an unauthorized person entering the restricted area are keep on increasing. In order restrict against these criminal activities, we need an automatic public security system. Each vehicle has their own Vehicle Identification Number ('VIN') as their primary identifier. The VIN is actually a License Number which states a legal license to participate in the public traffic. The proposed paper is to identify the vehicle with the help of vehicles License Plate (LP).LPRS is one the most important part of the Intelligent Transportation System (ITS) to locate the LP. In this paper certain existing algorithm drawbacks are overcome by the proposed morphological operations for LPRS. Morphological operation is chosen due to its higher efficiency, noise filter capacity, accuracy, exact localization of LP and speed.

Keywords
    Vehicle Identification Number (VIN), License Plate (LP), License Plate Recognition System (LPRS), Recognition, Morphological Operations.

## I.INTRODUCTION

LPR (License Plate Recognition) is an image-processing technology used to identify vehicles by their license plates. Intelligent transport system is a real-time, accurate, and efficient transportation management system can solve the various road problems generated by the traffic congestion, thus receiving more and more attention. The proposed work is mainly looking forward on number plate localization and detecting. The climatic conditions and type of plate for different countries vary. One algorithm could work very well for a given country's plate but very poorly for another. So based upon our requirement we should select the algorithms. This paper presents a new morphology based method for license plate location from car images. Morphological operations are used to fill the gaps between characters in edge image to make rectangular regions.

[1]

[1]**Prof.V.Karthikeyan** Department of Electronics and Communication Engineering SVS College of Engineering and Technology, Coimbatore-642109  E-mail:karthick77keyan@gmail.com
**Prof.K.Vijayalakshmi**, Assistant Professor, Department of Electrical and Electronics Engineering, SSK College of Engineering & Technology, Coimbatore-641105 E-Mail: vijik810@gmail.com

The algorithm uses morphological operations on the preprocessed, edge images of the vehicles. Characteristic features such as license plate width and height, character height and spacing are considered for defining structural elements for morphological operations.. Thresholding in a more formal way and extend it to techniques that are considerably more general. The basic morphological operations are erosion and dilation, dilation is used to fill the gaps or holes. We use a morphological operator for image of once dilated horizontally and the other time vertically. Another horizontal dilation is employed on the common bright pixels. The structuring elements of dilations are pixel horizontal or vertical lines. Due to digits and characters, a license plate contains many vertical edges. This feature is employed for locating the plate in an image. Many approaches have been proposed for edge detection. Sobel mask has a good performance compared with others; indeed, it is fast and simple. In general, there are two masks for Sobel, horizontal mask and vertical one. Closing also tends to smooth sections of contours but, as opposed to opening, it generally fuses narrow breaks and long thin gulfs, eliminates small holes, and fills gaps in the contour.

## II.LITERATURE REVIEW

The first objective of number plate recognition is to locate the number plate in the image. The purpose of this section is to identify possible candidate regions of the image in which the number plate might be contained.  The basic idea behind region growing is identifying the characteristics of the number plate such as color and then checking each pixel in the image for the identified characteristics. If a pixel is identified as containing the characteristics of a number plate pixel all its neighboring pixels are checked to see if they contain the characteristics of a number plate pixel.If a neighboring pixel contains the correct characteristics then we say that both pixels belong to the same region. This whole idea is recursively carried out until every pixel in the image has been examined. Region growing is a fast algorithm (i.e. O (n) algorithm) since each pixel is only examined once. Other advantages are that it extracts candidates with the correct shape, as it does not depend on the size of the region. Using the second method described in the paragraph above to make the identified regions be the same dimension or shape as a number plate makes region growing resistant to noise. A disadvantage of region growing is that since it is a recursive algorithm it requires a lot of memory usage. Setting the correct threshold for the preprocessing can also cause problems too when trying to create the binary image.

The method is based on adapting the variance of the Gaussian filter to the noise characteristics and the local variance of the image data. Based on observations of how the human eye perceives edges in different images, they concluded that in areas with sharp edges, the filter variance should be small to preserve the sharp edges and keep the distortion small. In smooth areas, the variance should be large so as to filter out noise. The major drawback of this algorithm is that it assumes the noise is Gaussian with known variance .In practical situations; however, the noise variance has to be estimated. The algorithm is also very computationally intensive.As it was mentioned, analyzing an image at different scales increases the accuracy and reliability of edge detection. Focusing on localized signal structures, Wavelet-based multi-resolution expansions provide compact representations of images with regions of low contrast separated by high-contrast edges. Additionally, the use of wavelets provides a way to estimate contrast value for edges on a space-varying basis in a local or global manner as needed.Wavelet transform (WT) is defined as the sum over the entire of rows and columns (i.e. spatial domain) of the image intensity function multiplied by scaled and shifted versions of the mother wavelet function. It results in coefficients that are function of the scale and shifts. Therefore, WT acts as a ―mathematical microscope, in which one can monitor different parts of an image by just adjusting focus on scale. An important property of WT is its ability to focus on localized structures, in this way, coarse and fine signal structures are simultaneously analyzed at different scales. Have no smoothing filter, and they are only based on a discrete differential operator. The earliest popular works in this category include the algorithms developed by Sobel (1970), Prewitt (1970), Kirsch (1971), Robinson (1977), and Frei-Chen (1977). They compute an estimation of gradient for the pixels, and look for local maxima to localize step edges. Typically, they are simple in computation and capable to detect the edges and their orientation, but due to lack of smoothing stage, they are very sensitive to noise and inaccurate.Though the classical methods were not in benefit of an independent smoothing module, they attempted to ease this drawback through the calculation of average over the image. Is the most known among the classical methods. The Sobel edge detector applies 2D spatial gradient convolution operation on an image. It uses the convolution masks to compute the gradient in two directions (i.e. row and column orientations), and then works out the pixels. Sobel edge detector is a simple and effective approach.The Canny operator was designed to be an optimal edge detector (according to particular criteria --- there are other detectors around that also claim to be optimal with respect to slightly different criteria). It takes as input a gray scale image, and produces as output an image showing the positions of tracked intensity discontinuities. The problem with Canny's edge detection is that his algorithm marks a point as an edge if its amplitude is larger than that of its neighbours without checking that the differences between this point and its neighbours are higher than what is expected for random noise.

### III.PROPOSED SYSTEM"

In order to overcome the drawbacks and to provide a efficient and accurate Localization of License Plate we proposed a License Plate Localization System using Morphological Operations.Identification, analysis, and description of the structure of the smallest unit of words Theory and technique for the analysis and processing of geometric structures – Based on set theory, lattice theory, topology, and random functions – Extract image components useful in the representation and description of region shape such as boundaries, skeletons, and convex hull.

RGB images have more depth it is difficult to process so we are converting the original image into gray scale image. Also image enhancement and edge detection are carried out on the gray-level image in order to adjust the structural property of the image in preparation for license plate region detection. In photography and computing, a grayscale digital image is an image in which the value of each pixel is a single sample, that is, it carries only intensity information. Images of this sort, also known as black-and-white, are composed exclusively of shades of gray, varying from black at the weakest intensity to white at the strongest. The input images are converted from 24-bit color image to 8-bit grayscale image using (1).

G (i, j) = 0.299 *Red + 0.587 *Green + 0.114 *Blue-   (1)

In image processing, to smooth a data set is to create an approximating function that attempts to capture important patterns in the data, while leaving out noise or other fine-scale structures/rapid phenomena. In smoothing, the data points of a signal are modified so individual points (presumably because of noise) are reduced, and points that are lower than the adjacent points are increased leading to a smoother signal. Smoothing may be used in two important ways that can aid in data stem analysis (1) by being able to extract more information from the data as long as the assumption of smoothing is reasonable and (2) by being able to provide analyses that are both flexible and robust.

The median filter is normally used to reduce noise in an image, somewhat like the mean filter. However, it often does a better job than the mean filter of preserving useful detail in the image.Like the mean filter, the median filter considers each pixel in the image in turn and looks at its nearby neighbors to decide whether or not it is representative of its surroundings. Instead of simply replacing the pixel value with the mean of neighboring pixel values, it replaces it with the median of those values.

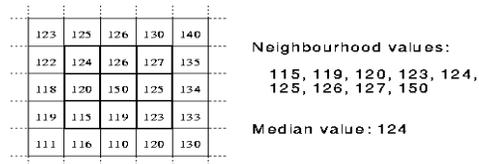

Fig.1 calculating the median value of a pixel neighborhood

The median is calculated by first sorting all the pixel values from the surrounding neighborhood into numerical order and then replacing the pixel being considered with the middle pixel value.(If the neighborhood under consideration contains an even number of pixels, the average of the two middle pixel values is used.) Calculating the median value of a pixel neighborhood as can be seen; the central pixel value of 150 is rather unrepresentative of the surrounding pixels and is replaced with the median value: 124. A 3×3 square neighborhood is used here larger neighborhoods will produce more severe smoothing. An image mask isolates parts of an image for processing. If a function has an image mask parameter, the function process or analysis depends on both the source image and the image mask. An image mask is an 8-bit binary image that is the same size as or smaller than the inspection image. Pixels in the image mask determine whether corresponding pixels in the inspection image are processed. If a pixel in the image mask has a nonzero value, the corresponding pixel in the inspection image is processed. If a pixel in the image mask has a value of 0, the corresponding pixel in the inspection image is not processed. Pixels in the source image are processed if corresponding pixels in the image mask have values other than zero. A mask affects the output of the function that inverts the pixel values in an image.

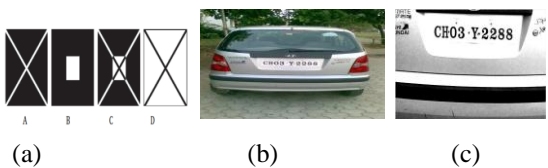

(a)  (b)  (c)

Fig.2 (a) Effect of image mask (b) &(c) Image after mask

Figure 2.ashows the inspection image. Figure (b) & (c) shows the image mask. Pixels in the mask with zero values are represented in black, and pixels with nonzero values are represented in white. Figure 3.3.5C shows the inverse of the inspection image using the image mask. Figure 2.a shows the inverse of the inspection image without the image mask. We can limit the area in which our function applies an image mask to the bounding rectangle of the region we want to process. This technique saves memory by limiting the image mask to only the part of the image containing significant information.Thresholding method is used to separate object and background, which is divided image into two modes .The way to resolve both categories is by assigning a thresholding value T. Each point (x, y) which have value f(x, y) > T is called point object, and each point (x, y) which have value f(x, y) ≤ T is called background object. A thresholded image g(x, y) is defined as:

g (x,y) = { Object if g (X, Y)
           Background if g (X, Y) ≤ T

T is a con t and is called global thresholding. Typically, an object pixel is given a value of "1" while a background pixel is given a value of "0." Finally, a binary image is created by coloring each pixel white or black, depending on a pixel's labels. The way to resolve both categories is by assigning a thresholding value T.
Binarization is a process where each pixel in an image is converted into one bit and you assign the value as '1' or '0' depending upon the mean value of all the pixel. If greater then mean value then its '1' otherwise its '0'.

## IV. SOBEL EDGE DETECTORS

The Sobel operator performs a 2-D spatial gradient measurement on an image and so emphasizes regions of high spatial frequency that correspond to edges. Typically it is used to find the approximate absolute gradient magnitude at each point in an input grayscale image.In theory at least; the operator consists of a pair of 3×3 convolution kernels as shown in Figure 3.3.8(1). One kernel is simply the other rotated by 90°. This is very similar to the Roberts Cross operator. These kernels are designed to respond maximally to edges running vertically and horizontally relative to the pixel grid, one kernel for each of the two perpendicular orientations. The kernels can be applied separately to the input image, to produce separate measurements of the gradient component in each orientation (call these Gx and Gy). These can then be combined together to find the absolute magnitude of the gradient at each point and the orientation of that gradient. The gradient magnitude is given by:

$$|G|=\sqrt{Gx^2 + Gx^2}$$

Typically, an approximate magnitude is computed using:
$$|G|= |G_x|+|G_y|$$

This is much faster to compute. The angle of orientation of the edge (relative to the pixel grid) giving rise to the spatial gradient is given as $\theta$ = arctan $(G_y/G_x)$.In this case, orientation 0 is taken to mean that the direction of maximum contrast from black to white runs from left to right on the image, and other angles are measured anti-clockwise from this. Often, this absolute magnitude is the only output the user sees --- the two components of the gradient are conveniently computed and added in a single pass over the input image using the pseudo-convolution operator shown in Figure 3.3.8(2).Using this kernel the approximate magnitude is given by:

$|G| = | (P_1+2×P_2+P_3)-(P_7+2×P_8+P_9)|+|(P_3+2 × P_6+P_9)-(P_1+2 × P_4+P_7)|$.

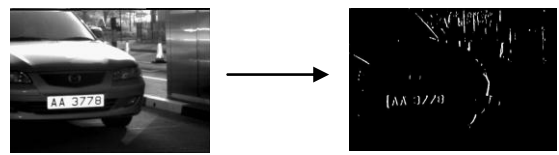

Fig.3   Edge detected Image

## V. CANNY EDGE DETECTOR

The Canny operator was designed to be an optimal edge detector (according to particular criteria --- there are other detectors around that also claim to be optimal with respect to slightly different criteria). It takes as input a gray scale image, and produces as output an image showing the positions of tracked intensity discontinuities. The Canny operator works in a multi-stage process. First of all the image is smoothed by Gaussian convolution. Then a simple

2-D first derivative operator (somewhat like the Roberts Cross) is applied to the smoothed image to highlight regions of the image with high first spatial derivatives. Edges give rise to ridges in the gradient magnitude image. The algorithm then tracks along the top of these ridges and sets to zero all pixels that are not actually on the ridge top so as to give a thin line in the output, a process known as non-maximal suppression.

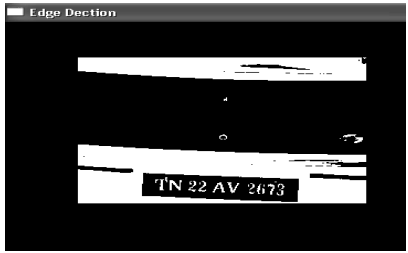

Fig .4 Plate Edge detection

The tracking process exhibits hysteresis controlled by two thresholds: T1 and T2, with T1 > T2. Tracking can only begin at a point on a ridge higher than T1.Tracking then continues in both directions out from that point until the height of the ridge falls below T2. This hysteresis helps to ensure that noisy edges are not broken up into multiple edge fragments.

## VI.MORPHOLOGICAL OPERATORS

Morphological operators often take a binary image and a structuring element as input and combine them using a set operator (intersection, union, inclusion, complement). They process objects in the input image based on characteristics of its shape, which are encoded in the structuring element. Usually, the structuring element is sized 3×3 and has its origin at the center pixel. It is shifted over the image and at each pixel of the image its elements are compared with the set of the underlying pixels. If the two sets of elements match the condition defined by the set operator (e.g. if the set of pixels in the structuring element is a subset of the underlying image pixels), the pixel underneath the origin of the structuring element is set to a pre-defined value (0 or 1 for binary images). A morphological operator is therefore defined by its structuring element and the applied set operator. For the basic morphological operators the structuring element contains only foreground pixels (i.e. ones) and `don't care's'. These operators, which are all a combination of erosion and dilation, are often used to select or suppress features of a certain shape, e.g. removing noise from images or selecting objects with a particular direction. The more sophisticated operators take zeros as well as ones and `don't care's' in the structuring element. The most general operator is the hit and miss, in fact, all the other morphological operators can be deduced from it. Its variations are often used to simplify the representation of objects in a (binary) image while preserving their structure. Morphological operators can also be applied to gray level images, e.g. to reduce noise or to brighten the image.

Dilation - grow image regions. Dilation is an operation that "grows" or "thickens" objects in a binary image. The specific manner and extent of this thickening is controlled by a shape referred to as a Structuring Element (SE).Computationally, structuring elements typically are represented by a matrix of 0's and 1's.Morphological dilation function sets the value of the output pixel to 1 if one of the elements in the neighborhood defined by the structuring element is one. Erosion - shrink image regions. Erosion is often used to remove irrelevant details from binary image. Erosion "shrinks" or "thins" objects in binary image. As dilation, the manner and extent of shrinking is controlled by a structuring element. Morphological erosion function sets the value of the output pixel to 1 if all of the elements in the neighborhood defined by the structuring elements are one

## VII.LICENSE PLATE DETECTION

The license plate localization is achieved by apply morphological operations .The statistical distribution of pixels value at edges of the image in the vertical direction is used to local possible license region . Rectangularity of image regions is found in order to detect license plate region. After obtaining the license plate region, the unwanted regions are cut off to obtain the license plate image, and then the image is thinned to one pixel wide. Locate license plate image bounding box. (i) Scan the binary image from top to bottom to obtain the image height. (ii) Scan the binary image from left to right to obtain the image width.(2) Centralization of the image. Calculate centre of gravity of the image using (1).

$$X = \frac{1}{N}\sum_{i=1}^{N} x(i)$$

Y = 1/N $\sum_{i=1}^{N}$ y (i)         (1)

## VIII.CONCLUSION

The proposed work presented a narrative effort of the exact Localization of License Plate. The proposed method narrates that RGB images have more difficult to process so we are converted the original image into gray scale image. Smoothing is used to extract more information from the data. The median filter is normally used to reduce noise in an image. An image mask isolates parts of an image for processing. Binary Thresholding method is used to separate object and background. Binarization is used to process of each pixel in an image is converted into one bit. The Sobel operator performs a 2-D spatial gradient measurement on an image .The canny operator takes as input a gray scale image, and produces as output an image showing the positions of tracked intensity discontinuities. Erosion is often used to remove irrelevant details from binary image. Structured removal of image region boundary pixels Opening is the combination of erosion-dilation. Structured filling of image region boundary pixels closing is the combination of dilation-erosion. Morphological methods were used to achieve the detection of License Plate. The localization and plate detection rate of the proposed work is so accurate and detection rate is very much prompt.

## IX. EXPERIMENTAL RESULT

In our proposed work the experimental results employ an essential task. Fig.5 shows the initial image is taken as original image Fig.6 shows the converted Filtered gray scale image Fig.7 illustrates the masked image which mask the whole image where the plate going to be located. Fig.8 demonstrates the Binary threshold Fig. 9 shows the binary mask image Fig. 10 illustrates the Binary processed image Fig. 11 point out the open image and Fig. 12 exemplify the boundary box image

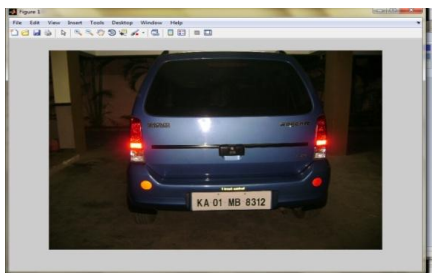

Fig .5  Input Image

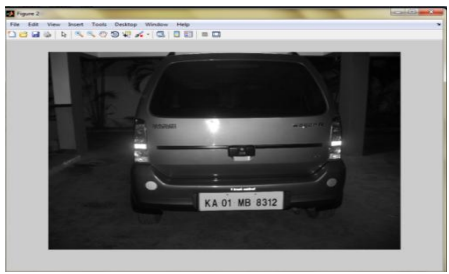

Fig. 6 Filtered Gray Scale Image

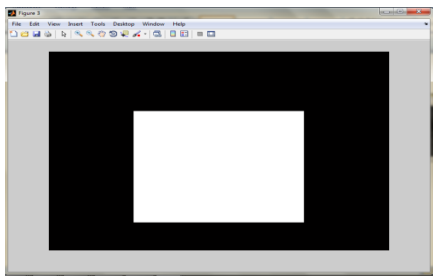

Fig .7 Mask

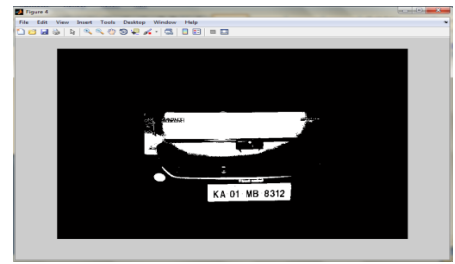

Fig. 8 Binary threshold Image

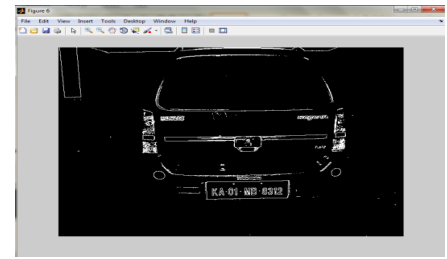

Fig .9 Binary-mask Image

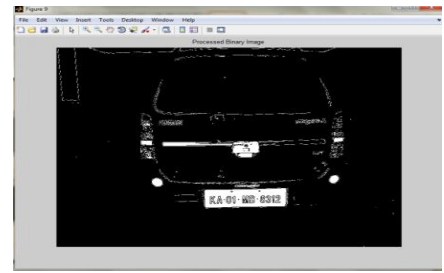

Fig.10 Binary Processed Image

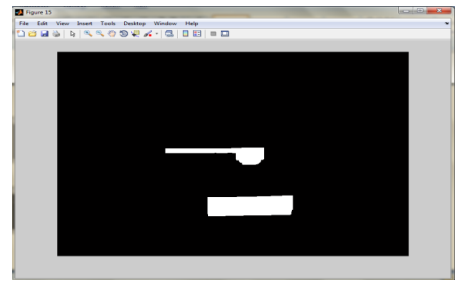

Fig .11 Open Image

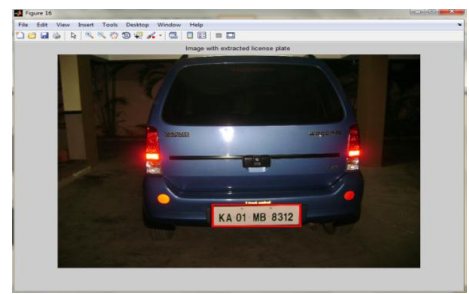

Fig .12 Boundary Box

## X. REFERENCES

1. SKohler, M. P., Mikkelsen, A. W. , Pedersen J. M., and Trangeled, M.,(2002), on 'Automatic recognition of license plates', Institute for Electronic System, Aalhorg University, Hontani, H., and Koga, T.,(2001),on 'Ch . Adebayo daramola1, e. adetiba1, a. u. adoghe1, j. a. badejo1, i. a samuel1 and t. fagorus (2011), on 'automatic vehicle identification system using license plate '.
2. Comelli, P., Ferragina, P., Granieri. M. N., and Stabile, F.(1995),on 'Optical recognition of motor vehicle license plates', IEEE Transactions on Vehicular Technology, vol. 44, no. 4, pp: 790-799,.
3. Cowell, J., and Hussein, F.(2002), on 'A fast recognition system for isolated Arabic characters', Proceedings Sixth International Conference on Information and Visualization, IEEE Computer Society, London, England, pp. 650-654, 2002.
4. Cowell, J., and Hussain, F.,(2001),on 'Extracting features from Arabic characters',Proceedings of the IASTED International Conference on Computer Graphics and Imaging, Honolulu, Hawaii, USA, pp. 201-206, 2001.
5. G. Deng and L.W. Cahill, (1994) ,on 'An adaptive Gaussian filter for noise reduction and edge detection',‖ in Proc. IEEE Nucl. Sci. Symp. Med. Im. Conf., 1994, pp. 1615–1619.
6. Hansen, H., Kristiansen, A. W., aracter extraction method without prior knowledge on size and information', Proceedings of the IEEE International Vehicle Electronics Conference (IVEC'01), pp. 67-72, 2001.
7. Hu, M. K., (1962) on 'Visual Pattern Recognition by Moment Invariant', IRE Transaction on Information Theory, vol IT- 8, pp. 179-187.
8. Jing-Ming Guo, Senior Member, IEEE, Yun-Fu Liu, Student Member, IEEE, and Chih-Hsien Hsia, Member, IEEE ,2012, on 'Multiple License Plates Recognition System'
9. Khalid Maglad, Dzulkifli Mohamad, Nureddin A. Abulgasem (2011), on 'Saudian Car License Plate Number Detection and Recognition Using Morphological Operation and RBF Neural Network'.
10. Kim, G. M., (1997), on 'The automatic recognition of the Plate of vehicle using the correlation coefficient and Hough Transform', Journal of Control, Automation and System Engineering, vol. 3, no.5, pp. 511-519, 1997
11. Salagado, L., Menendez, J. M., Rendon, E., and Garcia, N., (1999) on 'Automatic car plate detection and recognition Through Intelligent vision engineering', Proceedings of IEEE 33r Annual International Carnahan Conference on Security Technology, pp. 71-76, 1999
12. Park, S. FL, Kim, K. I., Jung, K., and Kim, H. J., (1999),on 'Locating car license plates using neural network', IEE Electronics Letters, vol.35, no. 17, pp. 1475-1477, 1999
13. Nieuwoudt, C, and van Heerden, R.,(1996), on 'Automatic number plate segmentation and recognition', Seventh annual South African workshop on Pattern Recognition, pp 88-93, IAPR,.
14. Naito, T., Tsukada, T., Yamada, K.s Kozuka, K., and Yamamoto, S., (2000),on 'Robust license-plate Recognition method for passing vehicles under outside Environment', IEEE Transactions on Vehicular Technology, vol: 49 Issue: 6, pp: 2309-2319, 2000.
15. Mitra Basu, ―Gaussian-Based Edge-Detection Methods(2002), on 'A Survey‖, IEEE Transactions On Systems, Man, And Cybernetics', Vol. 32, No. 3, August 2002, pp. 252-260 .
16. Lee, E. R., Earn, P. K., and Kim, H. J. (1994),on Automatic recognition of a car license plate using color image processing', IEEE International Conference on

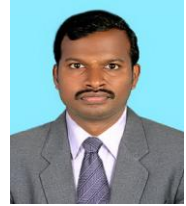

**Prof.V.Karthikeyan** has received his Bachelor's Degree in Electronics and Communication Engineering from PGP college of Engineering and technology in 2003 Namakkal, India. He received Masters Degree in Applied Electronics from KSR college of Technology, Erode in 2006. He is currently working as Assistant Professor in SVS College of Engineering and Technology, Coimbatore. He has about 7 years of teaching Experience.

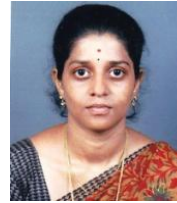

**Prof.K.Vijayalakshmi** has completed her Bachelor's Degree in Electrical & Electronics Engineering from Sri Ramakrishna Engineering College, Coimbatore, India. She finished her Masters Degree in Power Systems Engineering from Anna University of Technology, Coimbatore. She is currently working as Assistant Professor in Sri Krishna college of Engineering and Technology, Coimbatore. She has about 5 years of teaching Experience.